\documentclass[a4paper,twoside]{article}

\usepackage[utf8]{inputenc}
\usepackage{epsfig}
\usepackage{subfigure}
\usepackage{calc}
\usepackage{amssymb}
\usepackage{amstext}
\usepackage{amsmath}
\usepackage{amsthm}
\usepackage{multicol}
\usepackage{pslatex}
\usepackage{apalike}
\usepackage{SCITEPRESS}     

\begin{document}

\title{Field Of Interest Proposal for Augmented Mitotic Cell Count: Comparison of two Convolutional Networks}

\author{\authorname{Marc Aubreville\sup{1}, Christof A. Bertram\sup{2}, Robert Klopfleisch\sup{2} and Andreas Maier\sup{1}}
\affiliation{\sup{1}Pattern Recognition Lab, Computer Sciences, Friedrich-Alexander-Universität Erlangen-Nürnberg}
\affiliation{\sup{2}Institute of Veterinary Pathology, Freie Universität Berlin, Germany}
\email{marc.aubreville@fau.de}
}


\abstract{
Most tumor grading systems for human as for veterinary histopathology are based upon the absolute count of mitotic figures in a certain reference area of a histology slide. 
Since time for prognostication is limited in a diagnostic setting, the pathologist will oftentimes almost arbitrarily choose a certain field of interest assumed to have the highest mitotic activity. However, as mitotic figures are commonly very sparse on the slide and often have a patchy distribution, this poses a sampling problem which is known to be able to influence the tumor prognostication. On the other hand, automatic detection of mitotic figures can't yet be considered reliable enough for clinical application. In order to aid the work of the human expert and at the same time reduce variance in tumor grading, it is beneficial to assess the whole slide image (WSI) for the highest mitotic activity and use this as a reference region for human counting. \\
For this task, we compare two methods for region of interest proposal, both based on convolutional neural networks (CNN). For both approaches, the CNN performs a segmentation of the WSI to assess mitotic activity. The first method performs a segmentation of mitotic cells at the original image resolution, while the second approach performs a segmentation operation at a significantly reduced resolution, cutting down on processing complexity. \\
We evaluate the approach using a dataset of 32 completely annotated whole slide images of canine mast cell tumors, where 22 were used for training of the network and 10 for test. Our results indicate that, while the overall correlation to the ground truth mitotic activity is considerably higher ($0.936$ vs. $0.829$) for the approach based upon the fine resolution network, the field of interest choices are only marginally better. Both approaches propose fields of interest that contain a mitotic count in the upper quartile of respective slides.}

\onecolumn \maketitle \normalsize \vfill

\section{INTRODUCTION}
\label{sec:introduction}

Mitotic figures, i.e. cells undergoing cell division, are an important marker for tumor prognostication, as their density within tissue on a histology slide is assumed to be correlated with the proliferative rate of the tumor \cite{elston1991pathological}. Hence it is not surprising, that detection of mitotic figures has been the target of several object detection challenges in recent time \cite{Roux:2013kn,Veta:2015bi,veta2018predicting}. Detection of mitotic figures in digital whole slide images (WSI) is, however, not only a time-consuming task (as WSIs typically have very large image dimensions), but also a task presently not solved with a clinical applicable accuracy. This can be related to a number of factors: Firstly, the very definition of mitotic figures in histology slides is tricky, as their morphology is vaguely described as being without a nuclear membrane (post prophase) with \textit{hairy extensions} of nuclear material around the chromosomes \cite{VanDiest:1992hu}. Depending on factors such as inferior tissue quality often deriving from delayed tissue fixation, it is not always possible to unambiguously differentiate mitotic figures from mitotic-like structures such as pyknotic tumor cells of overstained nuclei. 
This leads to a high intra-observer variance \cite{PoulBoiesenParOlaBendahlLola:2009kv} in grading of cells between labs, schools and even individuals that are likely to reflect in data sets of mitotic figures applied for current developments of algorithms. Secondly, histology slides are subject to staining in order to make important details visible to the human eye. This dying procedure is however also subject to a number of influencing factors, including concentration and purity \cite{Horobin:1969fe} of coloring agents, slice thickness, the dying protocol and the dyed tissue itself. This leads to a significant color variance in hematoxylin and eosin stained tissue sections, which poses a challenge to pattern recognition methods, especially when color nuances may be a determining factor for cell classification. Lastly, mitotis is a sparse event in histology slides, which in turn leads to low numbers of events across databases and one can hypothesize, that not the complete biological variance spread is represented in current mitosis data sets like the Mitos \cite{Roux:2013kn} or TUPAC \cite{veta2018predicting} data set. 

Since mitotic figures can not be assumed as being evenly spread over the image, manual count within the usual diagnostic area of 10 consecutive High-Power-Fields (HPF, field of view at magnification of $400\times$) leads to an inherent \textbf{sampling problem}, as also assumed by Bonert and Tate \cite{Bonert:2017go}. It is thus strongly dependent on the actual region chosen intuitively by the pathologist, how many mitotic figures will be present within that area. Most grading schemes incorporate the mitotic count (MC, number of mitotic figures within 10 HPF) into the tumor grade, often using a direct thresholding approach. Especially for tumors with borderline MC around these thresholds, the area selection thus leads to a significant additional variance in the process of grading.

We assume that, in order to be clinical applicable, one interesting methodological approach would not be the direct recognition and fully automated count of mitotic figures in slides, as commonly performed, but rather the determination of a region of interest with a high mitotic figure density, assuming that this is also the region with highest proliferation. It is generally assumed that the region with highest proliferation has the strongest prognostic value for tumor grading \cite{Martin:1995uw,Baak:2008cm,Edmondson:2014bv}.

As such, the primary output of our approach will be the mitotic density of a given WSI. In order to do so, we depend on an intermediate mitotic figure segmentation map, which will be predicted by a deep convolutional network. In previous work, we have shown that the U-Net network architecture by Ronneberger \textit{et al.} \cite{Ronneberger:2015gk} is a very good candidate for this approach \cite{BVMpaper}. This model, however, comes with a quite high inherent complexity, and we wondered if a smaller version of the same approach directly targeting at a subsampled image map could yield similar overall performance.

\begin{figure*}[!ht]
  \centering
  \includegraphics[width=\linewidth]{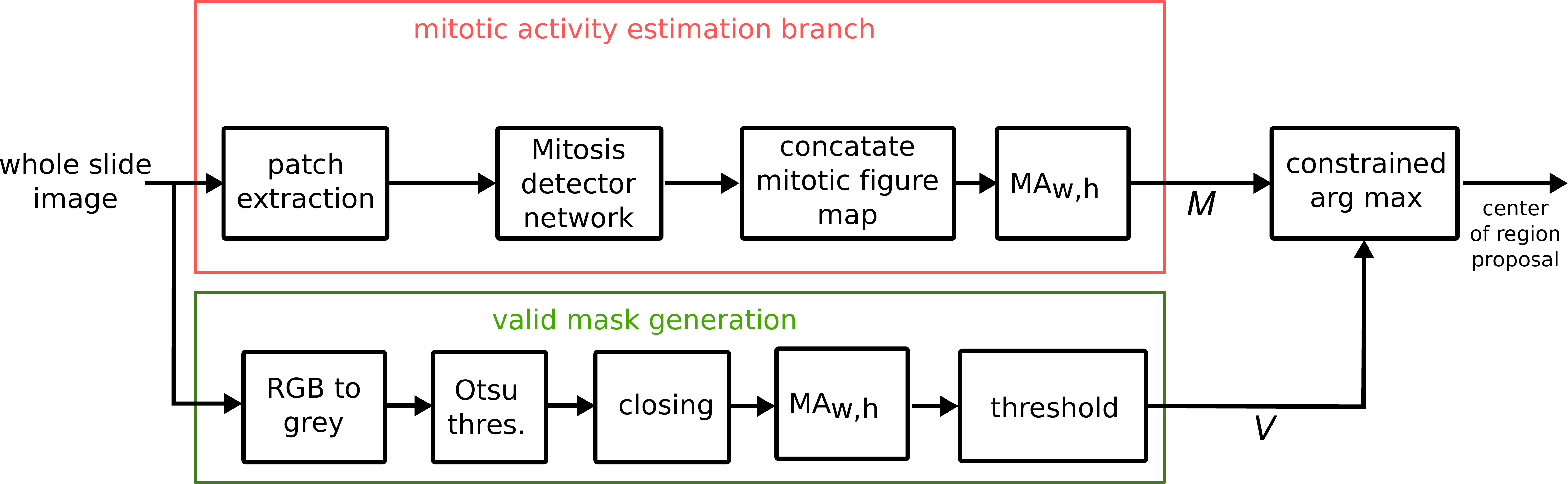}
  \caption{Overview of the proposed approach for mitotic count region proposal \cite{BVMpaper}. The upper path will derive singular mitotic annotations, followed by a moving average (MA) filter. The lower path derives an activity map of the image to exclude border regions of the image. This paper compares two mitosis detector CNNs, as further detailed in Fig. \ref{fig:cmdn} and Fig. \ref{fig:unet}}
  \label{fig:overview}
 \end{figure*}
\section{MATERIAL}

We annotated 32 whole slide images of canine cutaneous mast cell tumor, dyed with standard hematoxylin and eosin stain. All specimen was taken for routine tumor diagnostics, therefore no IRB approval was needed for this study. All slides were digitized using a linear scanner (Aperio ScanScope CS2, Leica Biosystems, Nussloch, Germany) at a magnification of $400 \times$, resulting in a digital resolution of $0.25$ microns per pixel. We used the open source software solution --- \cite{univis91884757} to attain a complete annotation map of all 32 WSI. The annotation process was performed in a partly computer-aided procedure, where the software would suggest partly overlapping segments of the whole slide image to the expert to annotate mitotic figures. In this process, we did not only annotate mitotic figures, but also granulocytes and other interesting cell types. It should be noted that also non-mitotic cells with similar appearance to mitotic figures were annotated with a designated class assigned to them. A second expert was asked to rate all cells blindly (i.e. not knowing the assigned class by the first expert). We only consider mitotic figures where both experts agreed on it being a mitosis for our data set, however, for hard negative examples, also mitotic figures annotated from one expert only or the aforementioned mitosis-like cells will be part of our training process. Following this definition of mitotic figure, our data set includes a total of 45,811 mitotic cells. To increase generalization, the data set purposefully includes cell tumors of different sizes and tumor grade, and thus the MC varies tremendously across cases.

\section{METHODS}

Mitosis detection is often considered an object detection approach, where singular events on an image have to be counted \cite[and others..]{Veta:2015bi,Li:2018ce,Ciresan:2013up}. This is due to the fact that mitotic events are often seen as a singular occurrence that can be described using a single (x,y) tuple. This is also reflected in several data sets such as MICCAI AMIDA 2013 \cite{Veta:2015bi}, ICPR MITOS-ATYPIA 2014 and TUPAC 16 \cite{veta2018predicting}, which use this for annotation. Other data sets, such as the Mitos 2012 data set \cite{Roux:2013kn}, provide segmentation information for mitotic cells, which is, however, a tedious process. In general, dataset creation for mitotic figure detection tasks, is a labour-intensive task, which might be one of the reasons for the limited data set size. To reduce the impact of this, we decided to use our own data set of canine mast cell tumors for this work. Additionally to an unprecedented size, our data set provides us with complete annotations of whole-slide-images, so border regions of the tumor as well as regions not containing tumor tissue will be included and enable an increased robustness of the approach.

\subsection{Field of Interest Proposal}

The goal of the algorithm is to suggest an area of the size of 10 adjacent High Power Fields with the highest mitotic count. Following Meuten \textit{et al.}, we assume this area to be a total of $2.37\,mm^2$ \cite{Meuten:2016jh}. We use an aspect ratio of $4:3$ for this rectangular selection. 

As depicted in Fig. \ref{fig:overview}, our approach consists of the generation of a map of mitotic figures on the WSI $M$ as well as a map of valid tissue $V$. For estimation of the mitotic count we utilize a convolutional neural network for generation of segmentation maps of mitotic figures. In order to retrieve the mitotic activity in a certain area, a moving average operator is used.

\begin{figure*}[!ht]
  \centering
  \includegraphics[width=\linewidth]{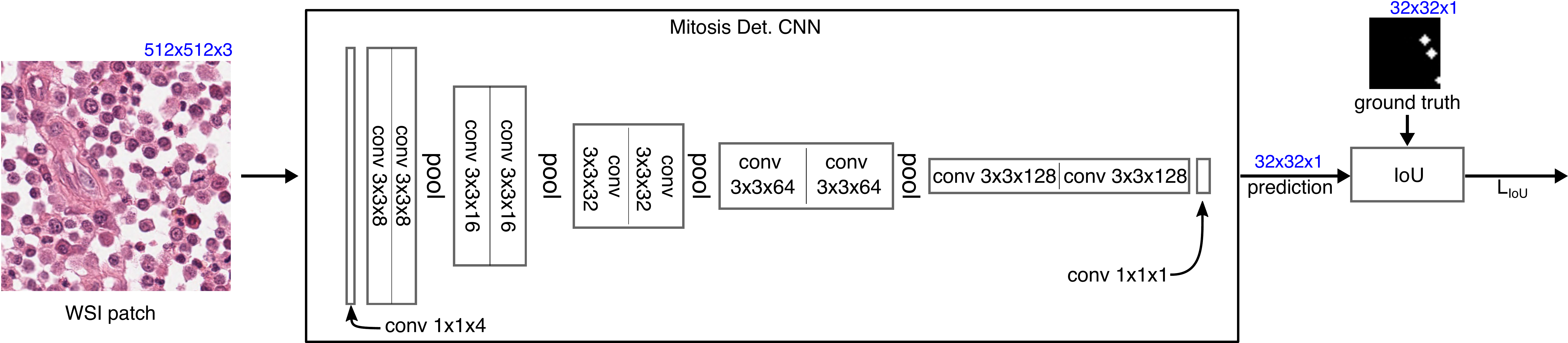}
  \caption{Overview of the coarse mitosis detection network (CMDN) and its training. The network predicts a 32$\times$32 map, i.e. a subsampling of 16, where mitotic figures are represented by filled circles. Intersection-over-Union (IoU) is used for optimization.}
  \label{fig:cmdn}
 \end{figure*}
 
 \begin{figure*}[ht!]
  \centering
  \includegraphics[width=\linewidth]{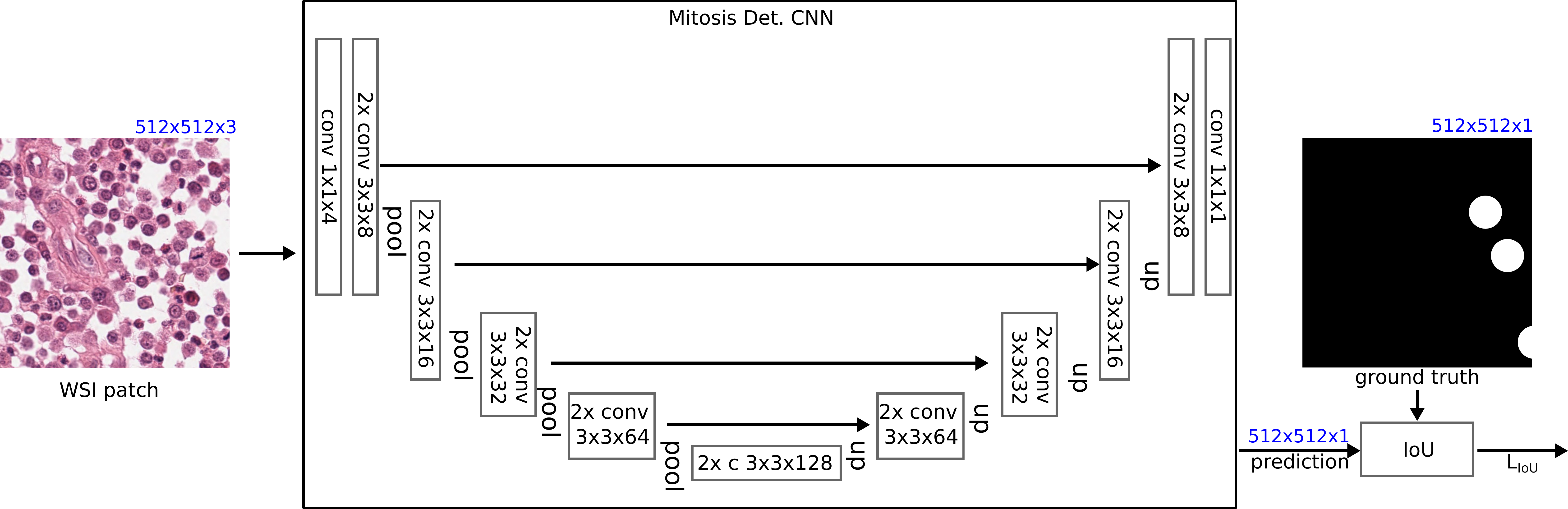}
  \caption{Overview of using U-Net as a mitosis detector. The network predicts a 512$\times$512 full-resolution map, where mitotic figures are represented by filled circles. Intersection-over-Union (IoU) is used for optimization.}
  \label{fig:unet}
 \end{figure*}
 
\subsubsection{Mitotic Activity Estimation}
For estimation of mitotic activity, the image is divided into overlapping (margin: 64\,px) patches with a size of $512\times 512$\,px. Due to the structure of the network, also other sizes would be applicable, reducing efforts for not covering the overlapping margins multiple times, but increasing memory footprint on the graphics card. The prediction of the network is being concatenated to yield an overall map $M$ of mitotic figure activity. 

\subsubsection{Valid Mask Estimation}

In order to exclude regions of the image that are partly uncovered by specimen, we construct a binary mask of tissue presence from the WSI at a low magnification. The image is converted to grey-scale, then a binary threshold is performed using Otsu's adaptive method \cite{Otsu:1979hf}. A closing operator is applied to reduce thin interruptions of the tissue map, and finally a moving average filter according to the size of the desired field of view (equivalent to 10 HPF) is being applied. Next, a thresholding with 0.95 is applied to retain only areas that are covered to at least 95\,\% with tissue, resulting in the valid mask $V$.

Lastly, both maps $M$ and $V$ are used to find the position of the maximum value, constrained to image areas where the valid mask is nonzero. We expect that these coordinates represent the center of ten high power fields with the highest mitotic count of the WSI.

\subsection{Comparison of Two Network Architectures}
Ronneberger's U-Net architecture \cite{Ronneberger:2015gk} has been successfully used in a large number of segmentation tasks throughout medical imaging, such as aortic stent segmentation \cite{Breininger:2018tr}, organ segmentation \cite{univis91881915} or bone and tumor segmentation \cite{kayalibay2017cnn}. We have shown previously \cite{BVMpaper}, that this architecture can also be used for direct mitotic figure segmentation. 

However, in this approach, we generate a fine (i.e. in the same resolution as the original image) segmentation map of the image, where a much more coarse version of the same map would be sufficient for the subsequent steps of forming a map of mitotic density estimates. We thus investigated the question if the downsampling path of the U-Net architecture might be sufficient for the given task, in effect removing the complete upsampling path with its skip connections and adding a simple 1$\times$1 convolution layer. We will denote this approach in the following as coarse mitosis detection network (CMDN). 

\subsubsection{Coarse Mitosis Detection Network}

The coarse network (see Fig. \ref{fig:cmdn}) consists of 5 stages of pairs of 2D convolution layers (filter kernel size: 3x3) followed by a maximum pooling operation (filter kernel size: 2x2) each. As in the approach by Ronneberger \textit{et al.}, the filter depth (or number of filter channels) doubles with each layer. A 1x1 convolution is being used at the input of the network for colour space adjustment, and another 1x1 convolution to generate the output mask with a dimension of $32\times32\times1$. Batch normalization and rectifying linear units (ReLU) as nonlinearities  are used after each convolutional layer. The final convolution layer uses a sigmoid activation function. As described in earlier works \cite{BVMpaper}, we utilize negative Intersection over Union (IoU) as a loss function for the task, and we minimize this using Adam Optimizer \cite{kingma2014adam} with Tensorflow. The use of IoU as a loss function, as proposed by Rahman and Wang \cite{rahman2016optimizing}, has the advantage of helping with the strong imbalance problem introduced by the sparsity of mitotic figures in WSI. The IOU operator is being applied on the network output and a ground truth estimate of mitotic figures (see Fig. \ref{fig:cmdn}). Here, since in the subsampled map, centers of mitotic figures will typically not be in the center of the sampling grid, we use sub-coordinate drawing of the mask (using the shift parameter of OpenCV's circle operation). This results in a more acurrate downsampled representation of the mitotic figure mask.

\subsubsection{U-Net as Mitosis Detection Network}

For comparison, we segment the same input images with a full-resolution mitotic figure map using Ronneberger's U-Net approach \cite{Ronneberger:2015gk} (see Fig. \ref{fig:unet}). We assume that, as in other evaluations, the skip-connections between the downsampling path and the respective same resolution of the upsamling path will help the network to find more accurate results. Admittedly, this network will have approximately twice the parameters of the original network, so it could potentially perform better due to its bigger capacity. 

\subsubsection{Training of the Networks}
Both networks have been trained for the exact same number of iterations. We observed that for both networks, the training had converged, as visible in a stable validation loss. For both networks, training samples were drawn randomly from the complete training set consisting of 22 Whole Slide Images. In these training images, the upper $80\,\%$ was used for training, while the lower $20\,\%$ was used for validation. 

We employed a strategy, where in each mini-batch of three images, one image would contain at least one mitotic figure, another would be drawn completely randomly, and one would be the hard example pick, containing at least one cell where either the experts did not agree on being a mitotic figure or it being classified as mitotic-figure-similar but not being a mitotic figure. Each of these images was taken as a crop with random rotation from the original WSI. For validation, images were drawn completely at random from the respective image region in order to be statistically as close as possible to the actual test set. Due to this approach of random sampling, we were not able to determine a training epoch as by the classical definition of the network having seen all training images once. Thus, we consider a run of 15,000 image iterations a pseudo-epoch. As our validation set is comparatively large, we chose a random pick of 6,000 images to be run after each epoch to evaluate the performance. Both networks have been trained for 150 pseudo-epochs.

\section{RESULTS}

\begin{figure}[!ht]
  \centering
  \includegraphics[width=\linewidth]{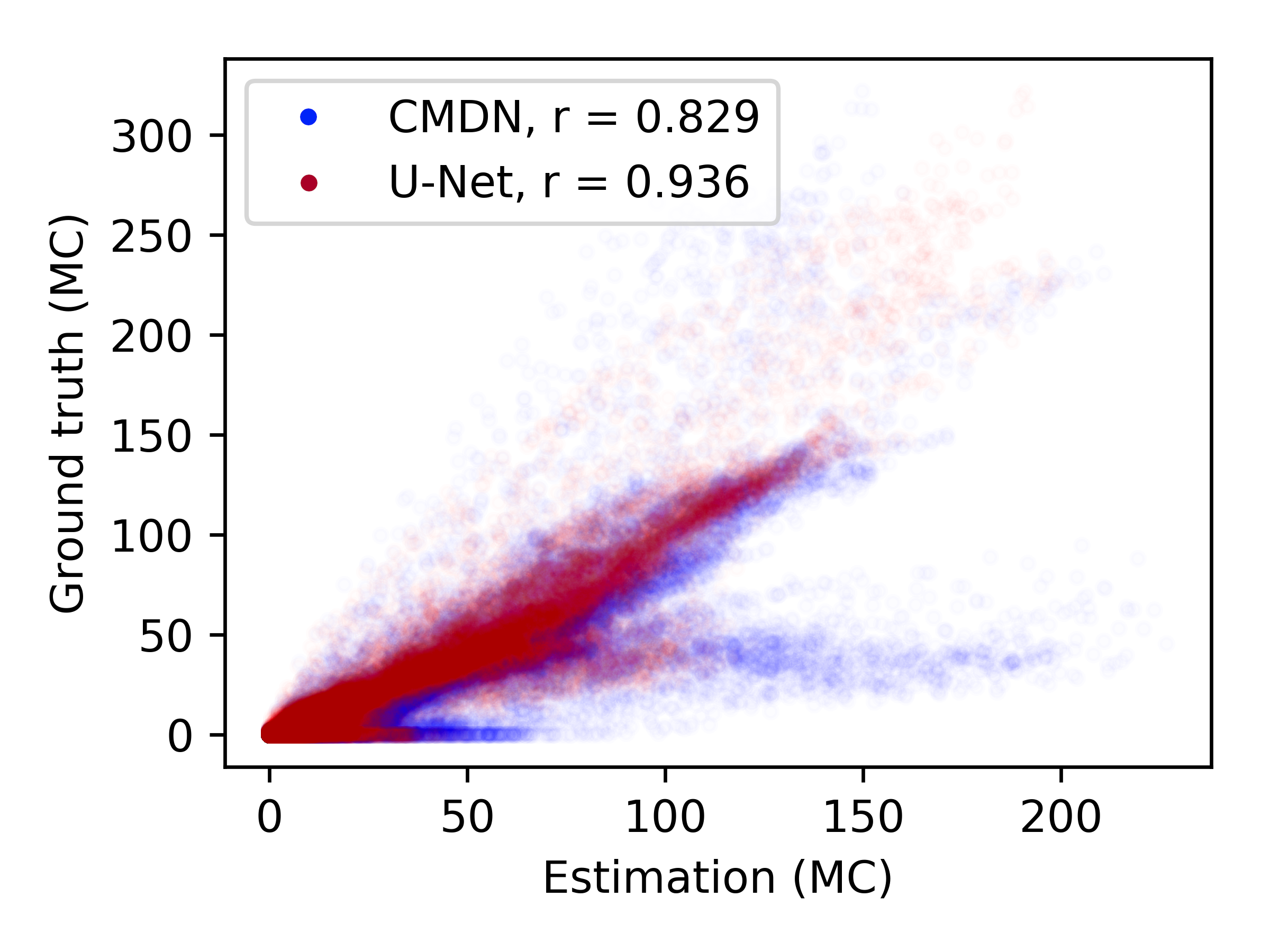}
  \caption{Relationship between ground truth mitotic count and estimate for approaches using the coarse network (CMDN, blue) and the U-Net (red) as mitotic figure detection network. Clearly, using the U-Net architecture leads to an overall better correlation to the ground truth mitotic count.}
  \label{fig:res_single}
 \end{figure}

Evaluating both approaches, we find a much higher correlation coefficient between the ground truth mitotic count map and the estimated map when using the U-Net architecture ($r=0.936$) compared to the coarse CMDN approach ($r=0.829$). As visible from Fig. \ref{fig:res_single} the CMDN had a tendency to overestimate mitotic activity in the slides. 

\begin{figure}[!ht]
  \centering
  \includegraphics[width=\linewidth]{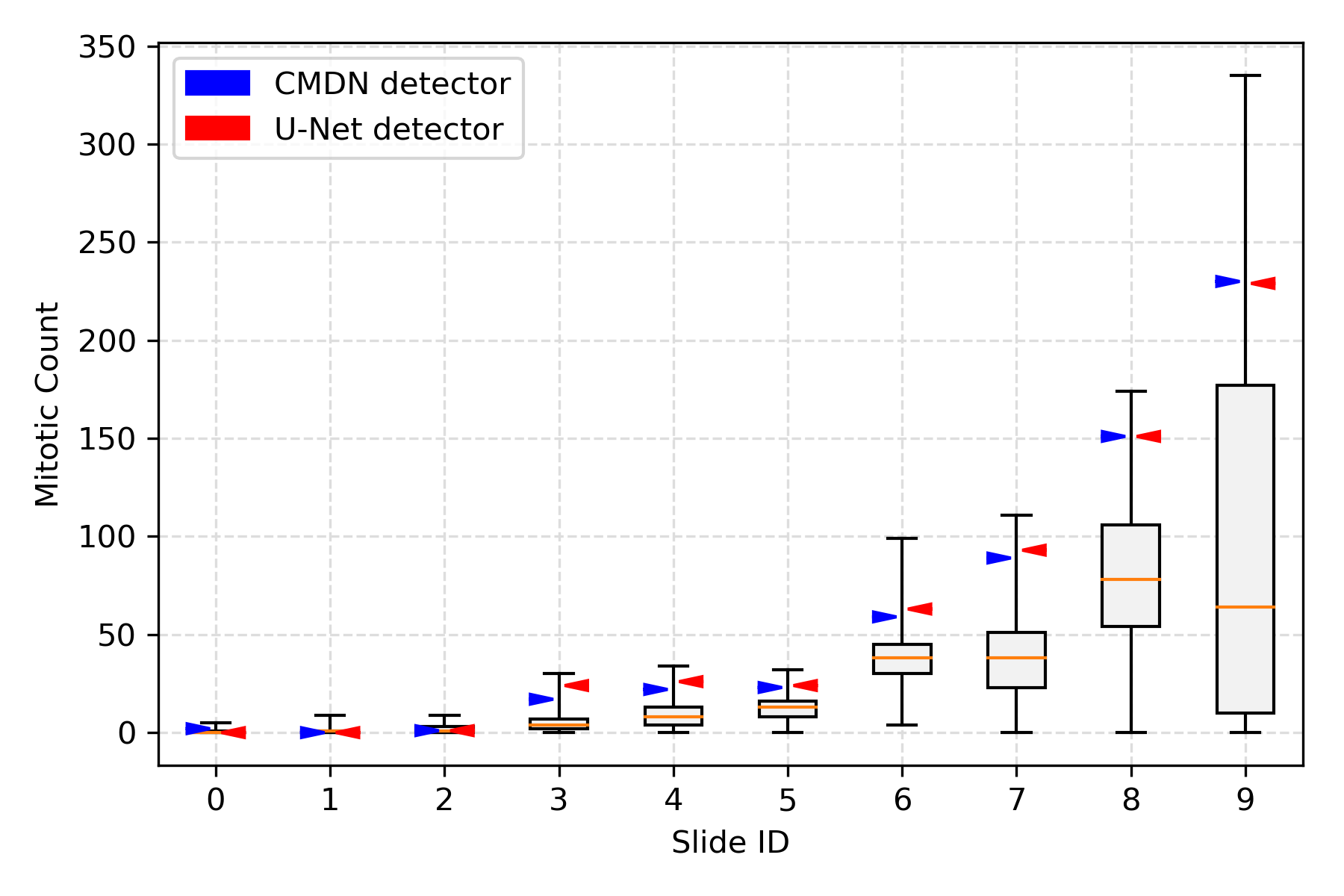}
  \caption{Box-whisker plots of mitotic count in all slides of our test data set. Only tumor tissue was included in this analysis. The arrows indicate the mitotic count (as of ground truth labels) of the proposed position by the U-Net detector (red) and the CMDN detector (blue) in the slide, i.e. the closer it is to the maximum value of the distribution, the better the estimate.}
  \label{fig:distribution}
 \end{figure}

\begin{figure*}[ht!]
	\centering
	\subfigure[Test slide 3]{
	\includegraphics[height=4cm]{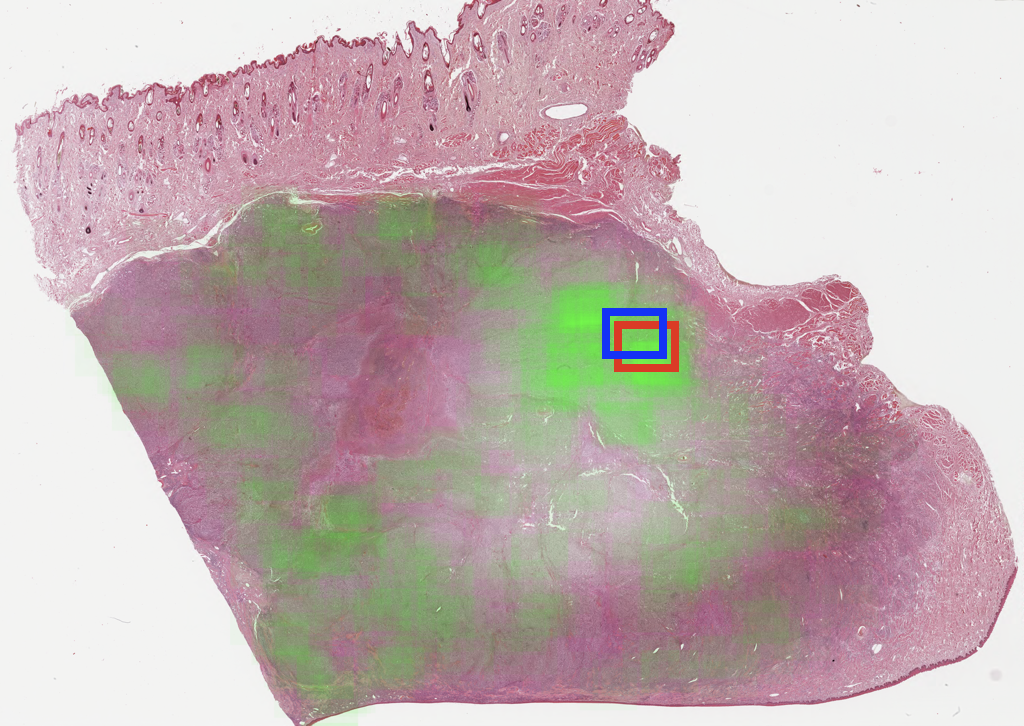}\label{fig:slide3}}
	\subfigure[Test slide 4]{
	\includegraphics[height=4cm]{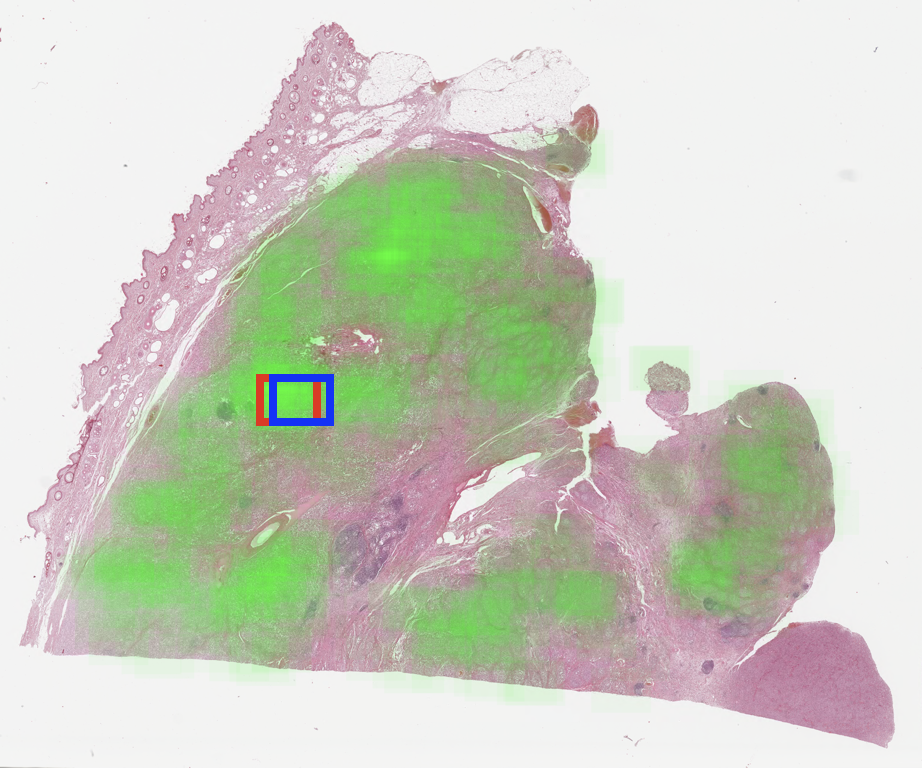}\label{fig:slide4}}
	\subfigure[Test slide 5]{
	\includegraphics[height=4cm]{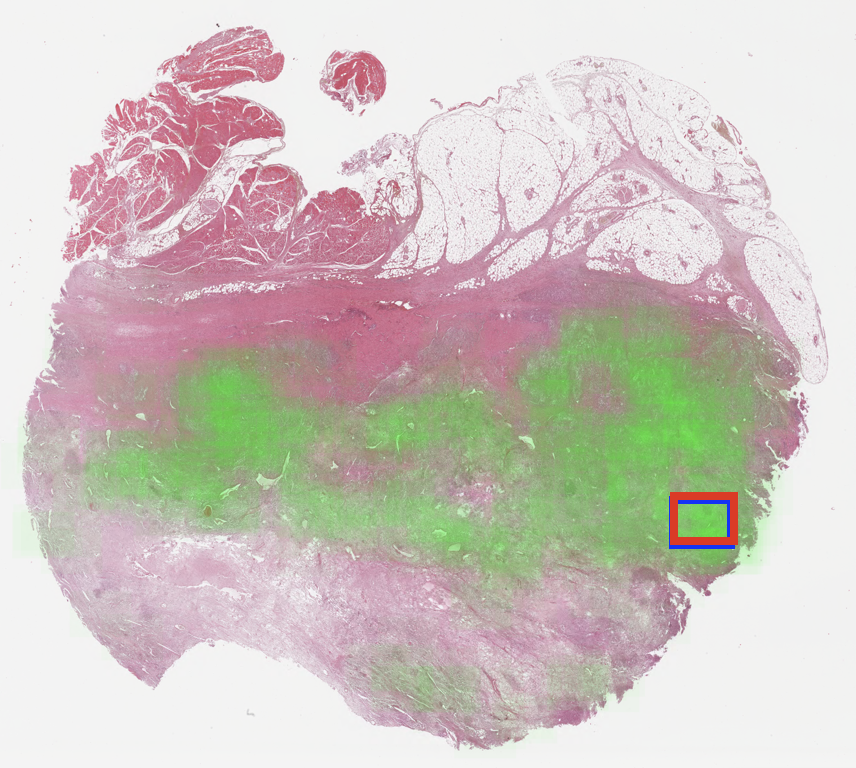}\label{fig:slide5}}
	\caption{Ground truth distribution of MC (green overlay, where higher opacity indicates higher MC) and region proposal of the approaches using U-Net (red) and CMDN (blue) for three slides of our test set. For all three slides, the proper choice of field of view has a high influence on the prognostication.}
	\label{fig:example}
\end{figure*}

This also reflects in an overall better performance in predicting a proper field of interest, as seen in Fig.~\ref{fig:distribution}. Although for most test slides, the differences were minor, we find a better region proposal for some slides, e.g. for test slide 3, which is a relevant borderline slide. As visible in Fig.~\ref{fig:slide3}, this slide has a rather unequal distribution of mitotic figures (and thus of MC) in the tissue. For all approaches, however, the position chosen in the relevant slides (3-9), yields a value in the upper quartile of mitotic count distribution (Fig.~\ref{fig:distribution}).

\begin{figure*}[ht!]
	\centering
	\includegraphics[width=\linewidth]{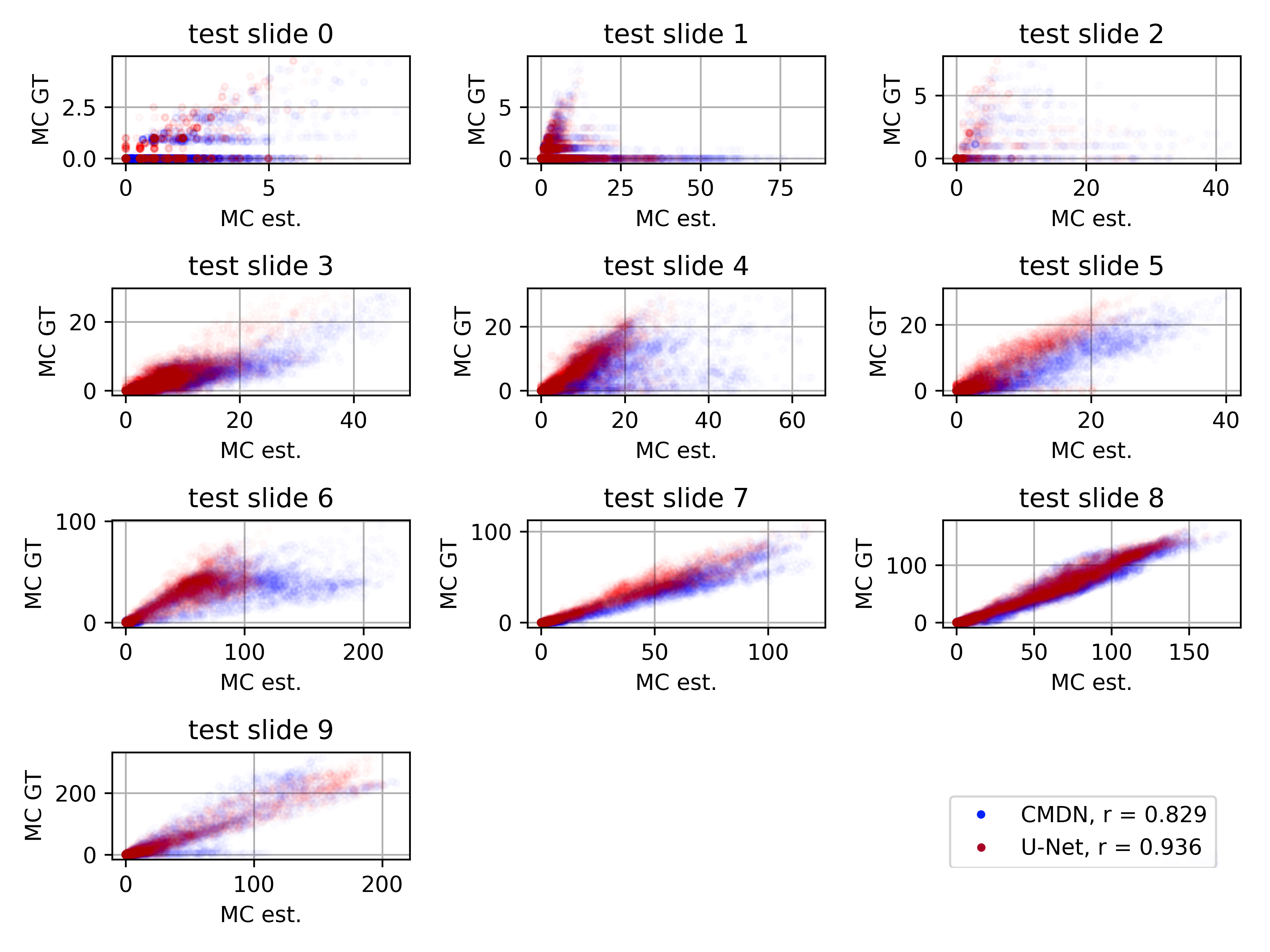}
	\caption{Semitransparent scatter plots of individual test slides with the U-Net detector in red and the CMDN detector in blue. As obvious from slides 3,4 and 5, the CMDN detector tends to overestimate the mitotic count especially in slides with borderline specimen.}
	\label{fig:correlations}
\end{figure*}

The evaluation of individual slides (Fig.~\ref{fig:correlations} and Table~\ref{tab:correlations}) shows, that correlation between mitotic count estimate and ground truth is rather weak for slides with very low mitotic activity (test slides 0 to 2). Here, both networks tend to overestimate the presence of mitotic figures. For borderline (3 to 5) and slides with high mitotic activity (6 to 9), the correlation is generally good. 

For all individual test slides, the proposed region reflects a region of high mitotic activity on the given WSI.

\begin{table*}
	\centering
	\begin{tabular}{l|cccccccccc}
		Test slide & 0 & 1 & 2 & 3 & 4 & 5 & 6 & 7 & 8 & 9\\
	\hline
		CMDN &  0.545 & 0.019 & 0.354 & 0.895 & 0.713 & 0.932 & 0.858 & 0.971 & 0.974 & 0.911 \\
U-Net &  0.617 & 0.111 & 0.322 & 0.848 & 0.872 & 0.919 & 0.924 & 0.976 & 0.983 & 0.973 \\
	\end{tabular}
	\caption{Correlation coefficients between estimated and ground truth mitotic count (MC) in the complete WSI for individual slides in the test set. While for many slides, performance between both approaches is comparable, slide 0, 4, 6 and 9 yield a clear advantage for the approach utilizing U-Net.}
	\label{tab:correlations}
\end{table*}

\section{DISCUSSION}
We demonstrated that, while the general problem of identifying  mitotic figures in whole slide images with high accuracy, is still far from being achieved, the outcomes of mitosis detection approaches might well serve as an intermediate step. In preselecting the field of interest containing the highest mitotic figure density, the algorithm can help the pathologist in determining the area of the tumor where the proliferation is the highest. Hence, we expect that such approaches can lead to a more reproducible grading and thus potentially better tailored treatment of the patient.  

The most crucial slides for the approach are slides 3 to 5, as also shown in Fig. \ref{fig:slide3} to \ref{fig:slide5}. Because the mitotic figure distribution in these slides is rather patchy i.e. with strong regional differences (see also Fig. \ref{fig:distribution} for absolute numbers) an arbitrary selection is likely to not yield the area with highest mitotic count, and thus the grading is subject to a possible strong variance. For all of these cases, both approaches were well able to pick an area with very high mitotic activity, with only minor differences in performance. The U-Net approach, while leading to a considerably higher correlation coefficient on the overall data set, did not lead to a significantly better overall performance.

Our approach did not employ stain normalization methods, as done in the majority of recent mitosis detection approaches \cite{veta2018predicting}. This was done in part, because the staining quality of our dataset is relatively stable due to the usage of a tissue stainer (ST5010 Autostainer XL, Leica, Germany) and all slides being created and scanned in the same lab. Additionally, we assume that with the high number of included WSI in the present study, natural variance of stain becomes less relavant. For application of this approach on another (possibly smaller) data set, however, we would recommend investigating a positive influence of such methods.

It is important to state that the results of this work were achieved on a limited test data set for canine mast cell tumors. While, theoretically, we would not assume different performance on different tumors, tissues or species, this should certainly be investigated. Another important question is to what degree the improved stability of region proposal, as shown in this work, would lead to a lower inter-rater-variability in grading, which we aim to deal with in future work.

 \newpage

\vfill
\bibliographystyle{apalike}

{\small
\begin{thebibliography}{}

\bibitem[Aubreville et~al., 2018a]{univis91884757}
Aubreville, M., Bertram, C., Klopfleisch, R., and Maier, A. (2018a).
\newblock {SlideRunner - A Tool for Massive Cell Annotations in Whole Slide
  Images}.
\newblock In Maier, A., Deserno, T.~M., Handels, H., Maier-Hein, K.~H., Palm,
  C., and Tolxdorff, T., editors, {\em {Bildverarbeitung f{\"u}r die Medizin
  2018 - Algorithmen - Systeme - Anwendungen. Proceedings des Workshops vom 11.
  bis 13. M{\"a}rz 2018 in Erlangen}}, pages 309--314.

\bibitem[Aubreville et~al., 2018b]{BVMpaper}
Aubreville, M., Bertram, C.~A., Klopfleisch, R., and Maier, A. (2018b).
\newblock Augmented mitotic cell count using field of interest proposal.
\newblock {\em arXiv preprint arXiv: arXiv:1810.00850}.

\bibitem[Baak et~al., 2008]{Baak:2008cm}
Baak, J. P.~A., Gudlaugsson, E., Skaland, I., Guo, L. H.~R., Klos, J., Lende,
  T.~H., S{\o}iland, H., Janssen, E. A.~M., and zur Hausen, A. (2008).
\newblock {Proliferation is the strongest prognosticator in node-negative
  breast cancer: significance, error sources, alternatives and comparison with
  molecular prognostic markers}.
\newblock {\em Breast Cancer Research and Treatment}, 115(2):241--254.

\bibitem[Boiesen et~al., 2000]{PoulBoiesenParOlaBendahlLola:2009kv}
Boiesen, P., Bendahl, P.~O., Anagnostaki, L., Domanski, H., Holm, E., Idvall,
  I., Johansson, S., Ljungberg, O., Ringberg, A., {\"O}stberg, G., and
  Fern{\"o}, M. (2000).
\newblock {Histologic grading in breast cancer: reproducibility between seven
  pathologic departments}.
\newblock {\em Acta Oncologica}, 39(1):41--45.

\bibitem[Bonert and Tate, 2017]{Bonert:2017go}
Bonert, M. and Tate, A.~J. (2017).
\newblock {Mitotic counts in breast cancer should be standardized with a
  uniform sample area}.
\newblock {\em BioMedical Engineering OnLine}, pages 1--8.

\bibitem[Breininger et~al., 2018]{Breininger:2018tr}
Breininger, K., Albarqouni, S., Kurzendorfer, T., Pfister, M., Kowarschik, M.,
  and Maier, A.~K. (2018).
\newblock {Intraoperative stent segmentation in X-ray fluoroscopy for
  endovascular aortic repair.}
\newblock {\em Int. J. Computer Assisted Radiology and Surgery}.

\bibitem[Chen et~al., 2018]{univis91881915}
Chen, S., Roth, H., Dorn, S., May, M., Cavallaro, A., Lell, M., Kachelrie{\"s},
  M., Oda, H., Mori, K., and Maier, A. (2018).
\newblock {Towards Automatic Abdominal Multi-Organ Segmentation in Dual Energy
  CT using Cascaded 3D Fully Convolutional Network}.
\newblock In Noo, F., editor, {\em {the fifth edition of The International
  Conference on Image Formation in X-ray Computed Tomography}}, pages 395--398.

\bibitem[Cire{\c s}an et~al., 2013]{Ciresan:2013up}
Cire{\c s}an, D.~C., Giusti, A., Gambardella, L.~M., and Schmidhuber, J.
  (2013).
\newblock {Mitosis detection in breast cancer histology images with deep neural
  networks.}
\newblock {\em Medical image computing and computer-assisted intervention :
  MICCAI ... International Conference on Medical Image Computing and
  Computer-Assisted Intervention}, 16(Pt 2):411--418.

\bibitem[Edmondson et~al., 2014]{Edmondson:2014bv}
Edmondson, E.~F., Hess, A.~M., and Powers, B.~E. (2014).
\newblock {Prognostic Significance of Histologic Features in Canine Renal Cell
  Carcinomas}.
\newblock {\em Veterinary Pathology}, 52(2):260--268.

\bibitem[Elston and Ellis, 1991]{elston1991pathological}
Elston, C.~W. and Ellis, I.~O. (1991).
\newblock Pathological prognostic factors in breast cancer. i. the value of
  histological grade in breast cancer: experience from a large study with
  long-term follow-up.
\newblock {\em Histopathology}, 19(5):403--410.

\bibitem[Horobin, 1969]{Horobin:1969fe}
Horobin, R.~W. (1969).
\newblock {The impurities of biological dyes: their detection, removal,
  occurrence and histological significance?a review}.
\newblock {\em The Histochemical Journal}, 1(3):231--265.

\bibitem[Kayalibay et~al., 2017]{kayalibay2017cnn}
Kayalibay, B., Jensen, G., and van~der Smagt, P. (2017).
\newblock Cnn-based segmentation of medical imaging data.
\newblock {\em arXiv preprint arXiv:1701.03056}.

\bibitem[Kingma and Ba, 2014]{kingma2014adam}
Kingma, D.~P. and Ba, J.~L. (2014).
\newblock Adam: Amethod for stochastic optimization.
\newblock In {\em Proc. 3rd Int. Conf. Learn. Representations}.

\bibitem[Li et~al., 2018]{Li:2018ce}
Li, C., Wang, X., Liu, W., and Latecki, L.~J. (2018).
\newblock {DeepMitosis: Mitosis detection via deep detection, verification and
  segmentation networks}.
\newblock {\em Medical Image Analysis}, 45:121--133.

\bibitem[Martin et~al., 1995]{Martin:1995uw}
Martin, A.~R., Weisenburger, D.~D., Chan, W.~C., Ruby, E.~I., Anderson, J.~R.,
  Vose, J.~M., Bierman, P.~J., Bast, M.~A., Daley, D.~T., and Armitage, J.~O.
  (1995).
\newblock {Prognostic value of cellular proliferation and histologic grade in
  follicular lymphoma}.
\newblock {\em Blood}, 85(12):3671--3678.

\bibitem[Meuten et~al., 2016]{Meuten:2016jh}
Meuten, D.~J., Moore, F.~M., and George, J.~W. (2016).
\newblock {Mitotic Count and the Field of View Area}.
\newblock {\em Veterinary Pathology}, 53(1):7--9.

\bibitem[Otsu, 1979]{Otsu:1979hf}
Otsu, N. (1979).
\newblock {A Threshold Selection Method from Gray-Level Histograms}.
\newblock {\em IEEE Transactions on Systems, Man, and Cybernetics},
  9(1):62--66.

\bibitem[Rahman and Wang, 2016]{rahman2016optimizing}
Rahman, M.~A. and Wang, Y. (2016).
\newblock Optimizing intersection-over-union in deep neural networks for image
  segmentation.
\newblock In {\em International Symposium on Visual Computing}, pages 234--244.
  Springer.

\bibitem[Ronneberger et~al., 2015]{Ronneberger:2015gk}
Ronneberger, O., Fischer, P., and Brox, T. (2015).
\newblock {U-Net - Convolutional Networks for Biomedical Image Segmentation.}
\newblock {\em MICCAI}, 9351(Chapter 28):234--241.

\bibitem[Roux et~al., 2013]{Roux:2013kn}
Roux, L., Racoceanu, D., Lom{\'e}nie, N., Kulikova, M., Irshad, H., Klossa, J.,
  Capron, F., Genestie, C., Le~Naour, G., and Gurcan, M.~N. (2013).
\newblock {Mitosis detection in breast cancer histological images An ICPR 2012
  contest.}
\newblock {\em Journal of pathology informatics}, 4:8.

\bibitem[Van~Diest et~al., 1992]{VanDiest:1992hu}
Van~Diest, P.~J., Baak, J. P.~A., Matze-Cok, P., Wisse-Brekelmans, E. C.~M.,
  van Galen, C.~M., Kurver, P. H.~J., Bellot, S.~M., Fijnheer, J., van Gorp, L.
  H.~M., Kwee, W.~S., Los, J., Peterse, J.~L., Ruitenberg, H.~M., Schapers, R.
  F.~M., Schipper, M. E.~I., Somsen, J.~G., Willig, A. W. P.~M., and Ariens,
  A.~T. (1992).
\newblock {Reproducibility of mitosis counting in 2,469 breast cancer
  specimens: Results from the Multicenter Morphometric Mammary Carcinoma
  Project}.
\newblock {\em Human Pathology}, 23(6):603--607.

\bibitem[Veta et~al., 2018]{veta2018predicting}
Veta, M., Heng, Y.~J., Stathonikos, N., Bejnordi, B.~E., Beca, F., Wollmann,
  T., Rohr, K., Shah, M.~A., Wang, D., Rousson, M., et~al. (2018).
\newblock Predicting breast tumor proliferation from whole-slide images: the
  tupac16 challenge.
\newblock {\em arXiv preprint arXiv:1807.08284}.

\bibitem[Veta et~al., 2015]{Veta:2015bi}
Veta, M., van Diest, P.~J., Willems, S.~M., Wang, H., Madabhushi, A., Cruz-Roa,
  A., Gonzalez, F., Larsen, A. B.~L., Vestergaard, J.~S., Dahl, A.~B., Cire{\c
  s}an, D.~C., Schmidhuber, J., Giusti, A., Gambardella, L.~M., Tek, F.~B.,
  Walter, T., Wang, C.-W., Kondo, S., Matuszewski, B.~J., Precioso, F., Snell,
  V., Kittler, J., de~Campos, T.~E., Khan, A.~M., Rajpoot, N.~M., Arkoumani,
  E., Lacle, M.~M., Viergever, M.~A., and Pluim, J. P.~W. (2015).
\newblock {Assessment of algorithms for mitosis detection in breast cancer
  histopathology images}.
\newblock {\em Medical Image Analysis}, 20(1):237--248.

\end{thebibliography}
}

\vfill
\end{document}